%% file: DOGlove.tex
\documentclass[conference]{IEEEtran}
\usepackage{times}

\usepackage[numbers]{natbib}
\usepackage{multicol}
\usepackage[bookmarks=true]{hyperref}

\usepackage{xcolor}
\usepackage{xspace}
\usepackage{graphicx}
\usepackage[font=footnotesize]{caption} 
\usepackage{amsmath}
\usepackage{bm}
\usepackage{booktabs}
\usepackage{pifont}
\usepackage{soul}

\renewcommand{\tablename}{Table} 

\newcommand{\oursystem}{DOGlove\xspace}

\newcommand{\scaletexttt}[1]{{\ttfamily\scalebox{0.95}[1]{#1}}}

\newcommand{\tick}{\ding{51}} 
\newcommand{\cross}{\ding{55}} 

\begin{document}

\title{\oursystem: Dexterous Manipulation with a Low-Cost Open-Source Haptic Force Feedback Glove}

\author{
    \authorblockN{Han Zhang$^{1}$, Songbo Hu$^{1}$, Zhecheng Yuan$^{1,2,3}$, Huazhe Xu$^{1,2,3}$}
    \authorblockA{$^{1}$ Tsinghua University, $^{2}$ Shanghai Qi Zhi Institute, $^{3}$ Shanghai AI Lab}
    \vspace{1mm}
    \authorblockA{\textbf{\textcolor{cyan}{\url{https://do-glove.github.io/}}}}
}


\IEEEpeerreviewmaketitle

\input{contents/0-abstract}
\input{contents/1-intro}
\input{contents/2-related-work}
\input{contents/3-glove-design-objectives}
\input{contents/4-hardware-system}
\input{contents/5-retargeting}
\input{contents/6-experiments}
\input{contents/7-limitations}
\input{contents/8-conclusion}

\input{contents/x-acknowledgment}

\bibliographystyle{plainnat}
\bibliography{references}

\end{document}

%% file: contents/0-abstract.tex
\input{captions/fg1-teaser}

\begin{abstract}
Dexterous hand teleoperation plays a pivotal role in enabling robots to achieve human-level manipulation dexterity. 
However, current teleoperation systems often rely on expensive equipment and lack multi-modal sensory feedback, restricting human operators' ability to perceive object properties and perform complex manipulation tasks.
To address these limitations, we present \oursystem, a low-cost, precise, and haptic force feedback glove system for teleoperation and manipulation.
\oursystem can be assembled in hours at a cost under 600 USD. It features a customized joint structure for 21-DoF motion capture, a compact cable-driven torque transmission mechanism for 5-DoF multidirectional force feedback, and a linear resonate actuator for 5-DoF fingertip haptic feedback.
Leveraging action and haptic force retargeting, \oursystem enables precise and immersive teleoperation of dexterous robotic hands, achieving high success rates in complex, contact-rich tasks. 
We further evaluate \oursystem in scenarios without visual feedback, demonstrating the critical role of haptic force feedback in task performance. In addition, we utilize the collected demonstrations to train imitation learning policies, highlighting the potential and effectiveness of \oursystem. 
\oursystem's hardware and software system will be fully open-sourced at \textcolor{cyan}{\url{https://do-glove.github.io/}}. 
\end{abstract}

%% file: captions/fg1-teaser.tex
\twocolumn[{%
\renewcommand\twocolumn[1][]{#1}%
\maketitle
\vspace{-5mm}
\includegraphics[width=\textwidth]{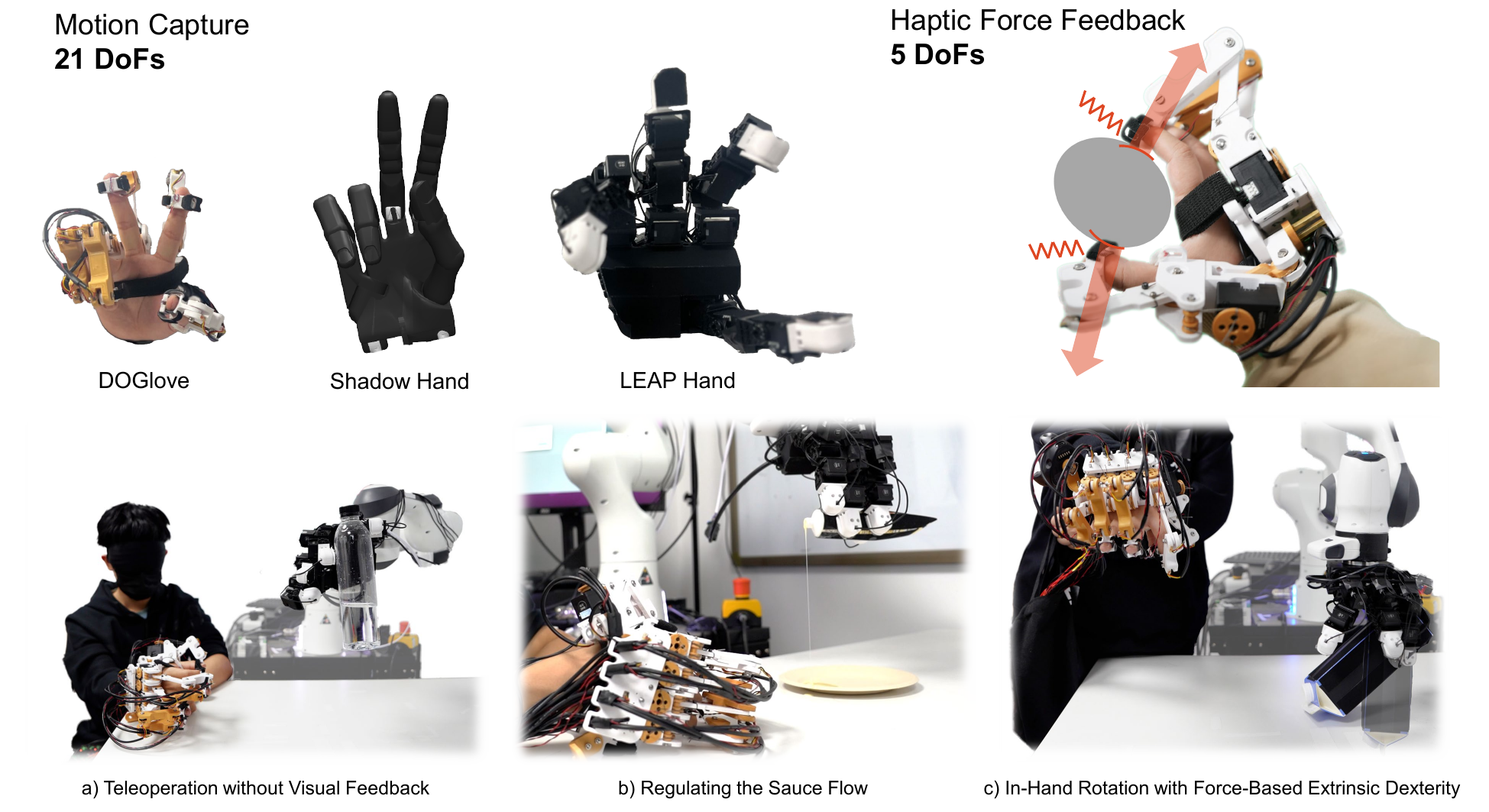}
\captionsetup{hypcap=false}
\captionof{figure}{\textbf{\oursystem}, a haptic force feedback glove designed for precise teleoperation and dexterous manipulation. It features 21-DoF motion capture and 5-DoF haptic force feedback. By leveraging action and force retargeting, it enables the teleoperation of dexterous hands for complex, contact-rich tasks, including:
a) without visual feedback, adjusting contact force with a bottle during teleoperation,
b) regulating the flow of condensed milk, and
c) performing in-hand rotation by using haptic force feedback to adjust friction.
}
\label{fig:teaser}
\vspace{0mm}
}]

%% file: contents/1-intro.tex
\input{captions/fg2-demo}
\section{Introduction}

Imitation learning~(IL) has shown significant promise in addressing complex manipulation tasks~\cite{chi2023diffusion,chi2024universal,zhao2023learning,Ze2024DP3}. However, it often necessitates a substantial amount of task-specific data to train a generalizable learning policy. Efficiently collecting and ensuring the high quality of such demonstrations remains a persistent and challenging problem for the robotic community.

Teleoperation is among the most commonly used methods for collecting demonstrations, often involving the development of a wide range of devices tailored to meet diverse data acquisition requirements. These devices enable the transfer of human manipulation behaviors to various robotic platforms~\cite{fu2024mobile, fang2024airexo,cheng2024open,ding2024bunny,iyer2024open}.
However, when it comes to dexterous hands, their high degrees of freedom~(DoFs) and inherent complexity impose even stricter demands on operational precision and the accuracy of human motion capture. Hence, it is crucial to design an intuitive, responsive, and highly precise device specifically suited for dexterous hand teleoperation applications.

Vision-based methods are primarily used for tracking the human hand in dexterous hand teleoperation. A simple approach involves using RGB cameras~\cite{qin2023anyteleop,ultraleap}, but the accuracy of hand gesture capture is often questioned and may be further limited by visual obstacles during hand-object interactions. Motion capture~(MoCap) systems~\cite{OptiTrack,Vicon,santaera2015low,wang2024dexcap} provide stable hand tracking. However, relying solely on visual feedback for teleoperation makes intuitive control challenging for the human operator.

Haptic force feedback offers additional perceptual characteristics beyond those provided by vision, such as the ability to sense an object's weight, friction, and softness. Integrating haptic feedback into teleoperation can enrich the feedback available during interaction and enable the completion of more challenging tasks. Recently, some commercialized force feedback gloves~\cite{dextarobotics, senseglove, manusmeta, haptx} have shown promise for enabling intuitive teleoperation. However, these solutions are often prohibitively expensive and require significant integration efforts to work with existing robot learning frameworks.

In this paper, we introduce \textbf{\oursystem}, a low-cost, fully open-sourced, and easy-to-manufacture haptic force feedback glove for dexterous manipulation. The glove can be assembled in hours for a total cost of 600 USD. Key features of \oursystem include:

\textbf{21-DoF motion capture:} 
\oursystem features an anthropomorphic design resembling the human hand, providing precise motion capture and a comfortable wearing experience. Moreover, we propose a customized joint structure that integrates a compact, low-cost yet accurate joint encoder, with the entire assembly measuring less than 15~mm in thickness.

\textbf{5-DoF haptic force feedback:} 
\oursystem leverages a cable-driven mechanism to deliver force feedback to each finger while maintaining a compact and cost-effective design. Additionally, each fingertip is also equipped with a linear resonant actuator~(LRA) to provide realistic haptic feedback. This integration of force and haptic feedback creates an immersive and responsive interface for dexterous manipulation.

\textbf{Action and haptic force retargeting:} We propose a general retargeting framework. For action retargeting, the rigid constraints of the glove allow fingertip positions to be mapped from the human hand to the target robotic hand. For haptic force retargeting, the combination strategy enables users to perceive contact information during teleoperation.

The resulting system, \oursystem, provides precise hand pose motion capture and the ability to sense interactions with manipulated objects. This enables human operators to intuitively and efficiently teleoperate dexterous hands. As shown in Fig.~\ref{fig:demo}, it further supports the completion of complex manipulation tasks.
We evaluate the necessity of haptic force feedback through a user study and further assess the teleoperation efficiency and data accuracy of \oursystem in several quantitative experiments.

Finally, we demonstrate that \oursystem seamlessly integrates with existing methods in robot learning. We use \oursystem to teleoperate the LEAP Hand mounted on a Franka robot arm, collecting data to train imitation learning policies. To foster further research, we will \textbf{open-source} the mechanical designs, circuit designs, embedded code, assembly instructions, URDF models, retargeting methods, and MuJoCo simulation environment at \textcolor{cyan}{\url{https://do-glove.github.io/}}.

%% file: captions/fg2-demo.tex
\begin{figure*}[t]
  \centering
  \includegraphics[width=\textwidth]{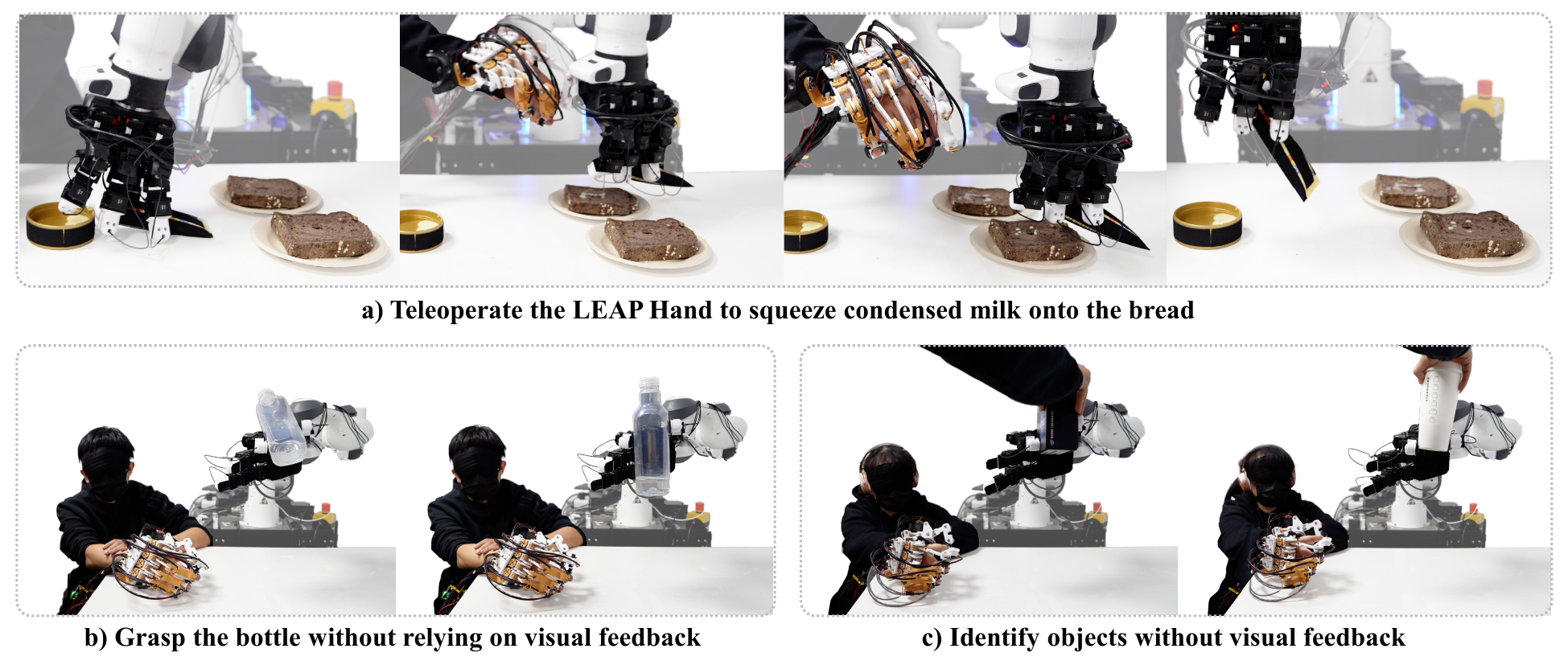}
  \caption{
    \textbf{Teleoperation demos.}
    a) While squeezing condensed milk, the operator regulates the flow using haptic force feedback from \oursystem.
    b) The operator grasps a slipping bottle without visual feedback.
    c) The user identifies object pairs solely through haptic force feedback.
  }
  \label{fig:demo}
  \vspace{-5mm}
\end{figure*}

%% file: contents/2-related-work.tex
\section{Related Work}
\subsection{Data Collection from Human Demonstrations}
A substantial amount of task-specific data is essential for imitation learning. In dexterous manipulation, obtaining high-quality hand motion data is critical for training effective policies.
Prior work includes extracting demonstration data from human videos~\cite{sivakumar2022robotic,xu2023xskill,xu2024flow,bharadhwaj2024towards} and hand trajectories~\cite{wang2023mimicplay,yang2022learning}. While these approaches are accessible and have shown promising results, the significant visual gap between recorded human demonstrations and the robot’s perception often makes real-world transfer challenging.
An alternative is using dedicated hardware for data collection to bridge this gap. Hand-held grippers~\cite{song2020grasping,chi2024universal,sanches2023scalable} have proven effective in capturing robot manipulation data. However, these systems are primarily designed for parallel grippers.
Another widely used approach is MoCap systems, which record human demonstrations and extract hand motion data. These systems include camera-based methods~\cite{qin2023anyteleop,zimmermann2019freihand}, glove-based tracking systems~\cite{wang2024dexcap,liu2017glove,liu2019high}, marker-based tracking~\cite{zhao2012combining}, and commercial MoCap solutions~\cite{taheri2020grab,fan2023arctic}. While MoCap offers high-precision tracking, bridging the embodiment gap between human and robotic hands remains a persistent challenge.

\subsection{Dexterous Hand Teleoperation} 
Collecting high-quality human demonstrations through robotic teleoperation systems~\cite{fang2024airexo,cheng2024open,ding2024bunny,iyer2024open} also plays a critical role for advancing dexterous manipulation.
Existing research has explored teleoperation from various perspectives, including leader-follower setups such as ALOHA~\cite{zhao2023learning,zhao2024aloha,fu2024mobile,aldaco2024aloha}.
However, teleoperating dexterous hands remains a significant challenge.
OpenTelevision~\cite{cheng2024open} leverages VR devices to capture hand poses and streams the pose information for retargeting to robotic hands. BiDex~\cite{shawbimanual}, on the other hand, implements a teleoperation system based on commercial motion capture gloves~\cite{manusmeta} and leader arms. 
Compared to these frameworks and other glove-based systems~\cite{liu2017glove,liu2019high}, \oursystem offers distinct advantages. It eliminates the need for expensive equipment while precisely capturing fingertip positions and delivering richer haptic force feedback to the operator. This system achieves accurate dexterous hand teleoperation with a low-cost setup, making it an efficient alternative.

\subsection{Teleoperation with Haptic Force Feedback}
While recent studies rely on visual information to capture environmental characteristics, vision alone inherently limits the richness of available sensory data. 
In contrast, haptic force feedback enhances the teleoperation experience by providing greater immersion and improving perception of the robot's status and movement compared to vision-based methods. 
Bunny-VisionPro~\cite{ding2024bunny} and Liu et al.~\cite{liu2019high} apply real-time haptic feedback to enable more accurate manipulation. 
Xu et al.~\cite{xu2025immersive} build a bilateral isomorphic bimanual telerobotic system using a commercial force feedback glove~\cite{dextarobotics} to enhance perception and improve performance in complex tasks.
NimbRo-Avatar~\cite{schwarz2021nimbro} and Mosbach et al.~\cite{mosbach2022accelerating} integrate commercial force feedback glove~\cite{senseglove} into dexterous teleoperation systems.
However, these approaches rely on specialized or expensive equipment. In contrast, \oursystem provides a highly accurate teleoperation system with integrated haptic force feedback at a significantly lower cost and can be widely used in dexterous manipulation.

%% file: contents/3-glove-design-objectives.tex
\input{captions/fg3-glove-explosion}
\section{Glove Design Objectives}

\oursystem is designed to precisely capture human hand poses and provide haptic force feedback for intuitive teleoperation. While ensuring these functionalities, the glove is optimized for accessibility by the research community, focusing on low cost, ease of manufacturing, and high performance. To achieve these goals, \oursystem incorporates the following design principles:

\subsection{\textbf{Low cost}}
Commercial products such as the SenseGlove Nova~\cite{senseglove} and Manus VR~\cite{manusmeta} cost more than 5,000~USD, making them prohibitively expensive for many researchers. In contrast, \oursystem provides a low-cost solution under 600~USD.

\subsection{\textbf{Ease of manufacturing}}
All parts of \oursystem are either readily available for purchase online or manufacturable using standard methods. The glove’s main body can be 3D-printed using a commodity 3D printer, while the remaining electronics and servos are easily sourced. The entire glove can be assembled within 6 hours.

\subsection{\textbf{Performance Sufficiency}}
To ensure precise fingertip position tracking, the glove's encoders deliver joint angle data with an error range of ±7.2°, which can be further minimized through careful calibration. For intuitive haptic force feedback, the servos provide sufficient stall torque to halt human finger movement, while the haptic engine supports multiple haptic waveforms to enhance tactile sensations.

\subsection{\textbf{Low latency}}
The MoCap system operates at a maximum frequency of 120~Hz, while the haptic force feedback system achieves a maximum frequency of 30~Hz. Together with the retargeting algorithm, the system ensures seamless operation at a minimum frequency of 30~Hz, providing a smooth and responsive teleoperation experience.

%% file: captions/fg3-glove-explosion.tex
\begin{figure*}[t]
  \centering
  \includegraphics[width=\textwidth]{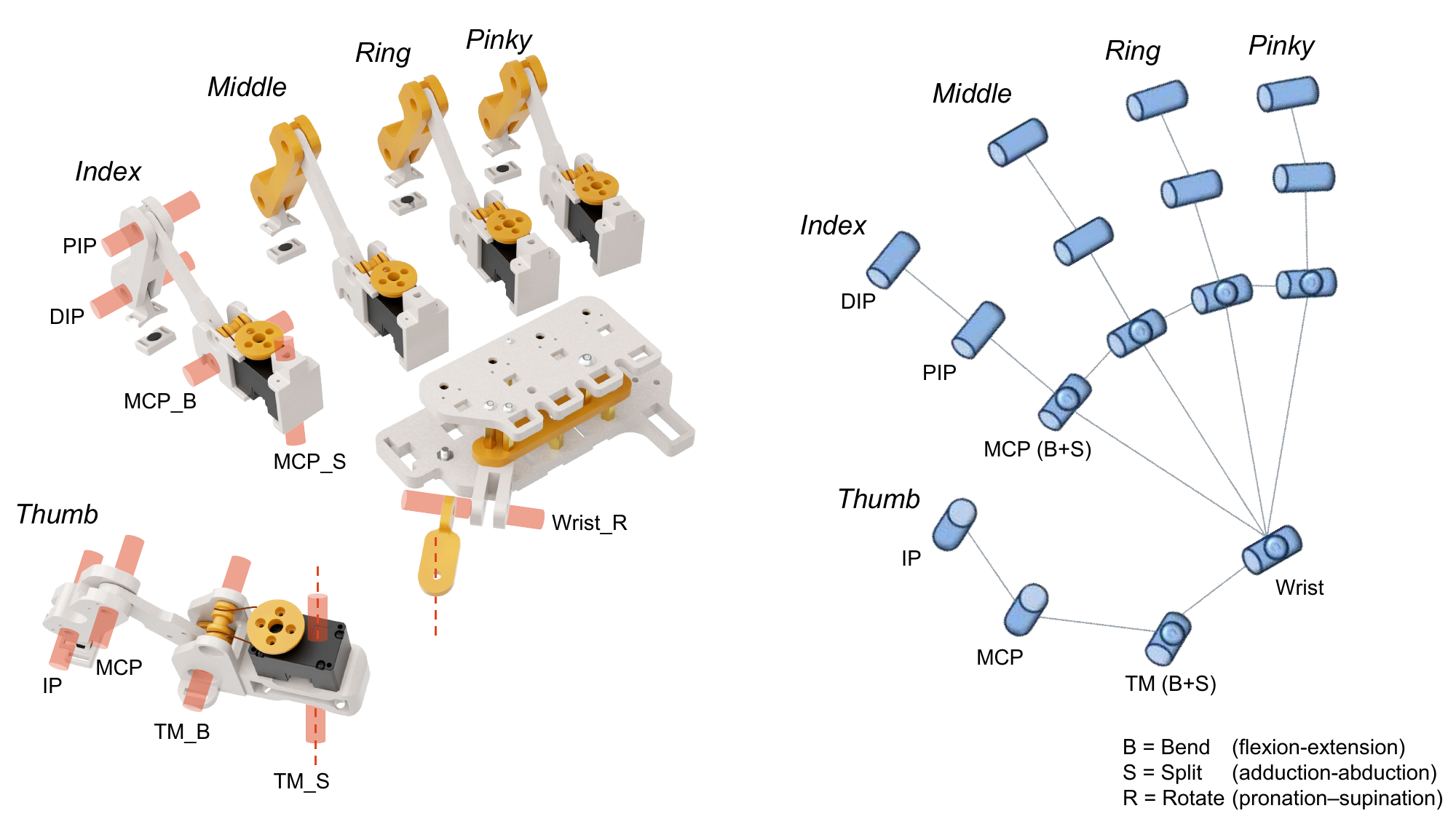}
  \caption{
  \textbf{The kinematic structure of \oursystem}, designed to replicate the kinematics of a human hand. 
  The MCP~(B+S) and TM~(B+S) joints are modeled as ball joints using a combination of two rotary joints. 
  The right figure from~\cite{cerulo2017teleoperation} illustrates the simplified human hand kinematics.
  }
  \label{fig:glove_explosion}
  \vspace{-3mm}
\end{figure*}

%% file: contents/4-hardware-system.tex
\section{Hardware System}

\subsection{Kinematic Design}

The kinematic design of \oursystem refers to the use of constraints to achieve desired movements, emulating the natural motion of a human hand. 
To ensure precise MoCap capabilities and a comfortable wearing experience, \oursystem are designed to closely resemble the anthropomorphic structure of the human hand.

Several studies~\cite{cerulo2017teleoperation, cerveri2007finger} model the hand skeleton as a kinematic chain, represented by a hierarchical structure of rigidly connected joints. As shown in Figure~\ref{fig:glove_explosion}, the kinematic structure of the human hand primarily consists of two types of joints: hinge joints and ball joints.

In the index, middle, ring, and pinky finger, distal interphalangeal~(DIP) and proximal interphalangeal~(PIP) joints are hinge joints with 1-DoF, allowing only flexion-extension movements. In contrast, the metacarpophalangeal~(MCP) joint is a ball joint with 2-DoF, allowing both flexion-extension and adduction-abduction movements. 

The thumb differs slightly in its structure. The interphalangeal~(IP) and metaphalangeal~(MCP) joints are hinge joints with 1 DoF, while the additional trapeziometacarpal~(TM) joint is a ball joint that supports both flexion-extension and adduction-abduction movements. To further enhance dexterity, an additional DoF at the wrist allows the thumb to perform pronation-supination movements.

To implement these joints in \oursystem, hinge joints are modeled as two linkages connected by a rotary joint, with a joint encoder installed on the rotary axis to capture motion. Ball joints are designed as a combination of two orthogonal rotary joints, each equipped with a joint encoder on its respective rotary axis.

\input{captions/fg4-improper-linkage}

The linkage lengths in \oursystem are designed to accommodate the majority of adult human sizes. To achieve this, a standard human finger length was first modeled, and the glove's linkage parameters were simulated to ensure an optimal range of motion. As shown in Figure~\ref{fig:improper_linkage}, improper linkage lengths can obstruct the natural flexion-extension of the fingers, leading to discomfort and reduced MoCap performance. Furthermore, \oursystem features a modular design where all fingers share a common structural framework. This modularity enables users to replace linkages with customized sizes as needed, enhancing both adaptability and usability.

\subsection{Finger Design}

\input{captions/fg6-joint_explosion}

As shown in the exploded-view in Figure~\ref{fig:glove_explosion}, \oursystem is composed of the thumb, index, middle, ring, and pinky finger assemblies, along with the palm base structure. The design of each finger assembly follows a modular approach, ensuring consistent structural elements across all fingers.

The exploded view of a single finger is illustrated in Figure~\ref{fig:joint_explosion}. The highlighted area indicates the basic components of a rotary joint. Each rotary joint is constructed using an M4×15 shoulder screw to connect the finger linkages, ball bearing, and joint encoder, secured with an M3 locknut. This design ensures smooth and reliable joint rotation.
The main body of the finger, colored white and gold, is 3D printed using PETG material for ease of fabrication and durability.

Given the limited space on the back of the human hand, the finger assembly's width is constrained to less than 26 mm. Simultaneously, to provide effective force feedback, the actuator must deliver a stall torque of at least 0.5 N·m. Additionally, adjustable stiffness requires the actuator's current to be regulated. Since the actuator is directly connected to the pulley system as a rotary joint for $MCP_B$, it is essential to measure its rotary position in real time to achieve precise joint angle control.
Considering these design requirements, the \texttt{Dynamixel XC/XL330} servo motors were selected as the actuators for force feedback. It fulfills the torque, size, and real-time position measurement needs, making it a suitable choice for \oursystem.

\vspace{1mm}
\subsubsection{\textbf{Joint Encoders}}
\leavevmode

To integrate joint encoders into the finger linkages, the encoders must be compact while maintaining high precision. Additionally, as 16 encoders are required in combination with 5 servo motors to achieve 21-DoF MoCap capabilities, the cost of each encoder needs to be affordable. Considering these constraints, we selected the \texttt{Alps RDC506018A} rotary sensor as the joint encoder. 
This compact encoder (W11~mm $\times$ L14.9~mm $\times$ H2.2~mm) is easily integrated into the 3D-printed joint structures. The encoder operates as a variable resistor, changing its resistance as the shaft rotates. 

The resistance changes are converted into voltage signals using a simple voltage divider circuit. Due to the encoder's linear response, the voltage output is proportional to the actual joint angle. These voltage signals are read by an Analog-Digital Converter (ADC) module. For precise conversion, we use the \texttt{TI ADS1256}, a low-noise 24-bit ADC operating at 30k samples per second. The converted signals are then sent to a microcontroller unit~(MCU), the \texttt{ST Electronics STM32F042K6T6}, which operates at a clock speed of 48 MHz. To optimize system performance and reduce OS scheduling overhead, the STM32's Direct Memory Access~(DMA) feature is utilized to accelerate joint encoder readings. Finally, the processed joint data is transmitted to the host machine via a serial port on the STM32.

The voltage readings are mapped directly to joint angles under the assumption that the supply voltage of the STM32~(approximately 3.3~V) corresponds to 360°, while the ground voltage~(0~V) corresponds to 0°. Using the ADC output voltage, the joint angle is calculated as:
\begin{equation}
    \alpha_{\text{joint}} = \frac{\text{V}_{\text{ADC}}}{\text{V}_{\text{CC}}} \cdot 360
    \label{eq:joint_angle}
\end{equation}

The primary error in this conversion comes from the linearity error of the encoder which is ±2\% according to its datasheet. This results in an angular error of ±7.2° when measuring joint angles. To mitigate this, we employ a calibration process. Using an external high-precision joint encoder, we map the voltage reading to an actual joint angle, creating a correction table for each encoder. With this calibration, the error can be reduced to within ±1°.

\vspace{2mm}
\subsubsection{\textbf{Cable-Driven Force Feedback Structure}}
\leavevmode

To provide force feedback on the human fingers, the output torque of the Dynamixel servo must be transmitted to the glove's finger linkage system. As illustrated in Figure~\ref{fig:joint_explosion}, the rotary axis of the servo and the rotary axis of the $MCP_B$ joint are misaligned. Consequently, a transmission mechanism is required to transfer the torque effectively.

Although a bevel gear system could serve as a potential solution, its implementation would require significant space to accommodate the two orthogonal gears. Additionally, transmitting large torque through gears can cause deformation in the gear shaft, leading to gear slippage. In contrast, a cable-driven mechanism offers a more compact design while ensuring stable torque transmission.

Traditional cable-driven systems typically provide unidirectional force transmission on the tension side, relying on a spring to generate force in the opposite direction. However, this approach introduces unrealistic feedback sensations. While using two servos per finger could resolve this issue, it would significantly increase the glove's weight and cost. 

\input{captions/fg7-pulley-system}

To address these challenges, \oursystem utilizes a pulley system to provide the bi-directional force feedback, as shown in Figure~\ref{fig:pulley_system}.
\oursystem uses a 0.6~mm stainless steel braided wire as the cable, chosen for its strength and durability. The \textit{Servo Pulley} connects the servo to the finger linkage~(\textit{Finger Middle}) via the \textit{Finger Pulley}, maintaining a 1:1 transmission ratio. To minimize friction during transmission, the \textit{Fixed Pulley} is used to redirect the cable’s path.
When the \textit{Servo Pulley} rotates clockwise, the tension on \textit{Cable B} increases, causing \textit{Finger Pulley} to rotate clockwise. The extra slack on the \textit{Cable A} side is taken up by the \textit{Servo Pulley A}. Since the finger linkage is fixed to the \textit{Finger Pulley}, it also rotates clockwise, resulting in the extension movement of the $MCP_B$ joint. Similarly, when the \textit{Servo Pulley} rotates counterclockwise, the tension shifts to \textit{Cable A}, producing a flexion movement of the $MCP_B$ joint.

This configuration enables bi-directional torque transmission while maintaining a simple, compact, and cost-effective design.

\vspace{2mm}
\subsubsection{\textbf{Fingertip Haptic Feedback}}
\leavevmode

To further enhance the operator’s tactile experience, each fingertip in \oursystem is equipped with a tactile actuator. 

Traditional haptic actuators include eccentric rotating mass~(ERM) motors and linear resonate actuators~(LRAs). Limited by the inertia of the rotating mass, ERM motors are slow to start and stop, making it challenging to produce complex waveforms needed for subtle tactile sensations. On the contrary, LRAs offer linear motion, resulting in a cleaner and more precise tactile output.

In \oursystem, we use LRAs with a diameter of 8~mm and a height of 2.5~mm, installed close to the fingertips. These LRAs provide vibration stimuli by resonating at approximately 240~Hz along Z axis, which is orthogonal to the fingertip surface. Operating at 1.2~$V_\mathrm{rms}$, the LRAs generate high-quality haptic waveforms. To fully leverage the potential of the LRA, we employ the \texttt{TI DRV2605L} motor driver, which includes the licensed \texttt{Immersion TouchSense\textsuperscript{\textcopyright} 2200} haptic library. This driver supports over 100 pre-programmed waveforms, allowing \oursystem to deliver realistic and refined haptic feedback.

\subsection{Wrist Localization}

In \oursystem, we design a shell with a 1/4 inch screw connector to accommodate external wrist localization devices. For our experiments, we use the \texttt{HTC Vive Tracker} for real-time wrist position tracking. However, the design is compatible with other solutions, depending on the user's requirement.

%% file: captions/fg4-improper-linkage.tex
\begin{figure}[ht]
  \centering
  \vspace{2mm}
  \includegraphics[width=0.92\linewidth]{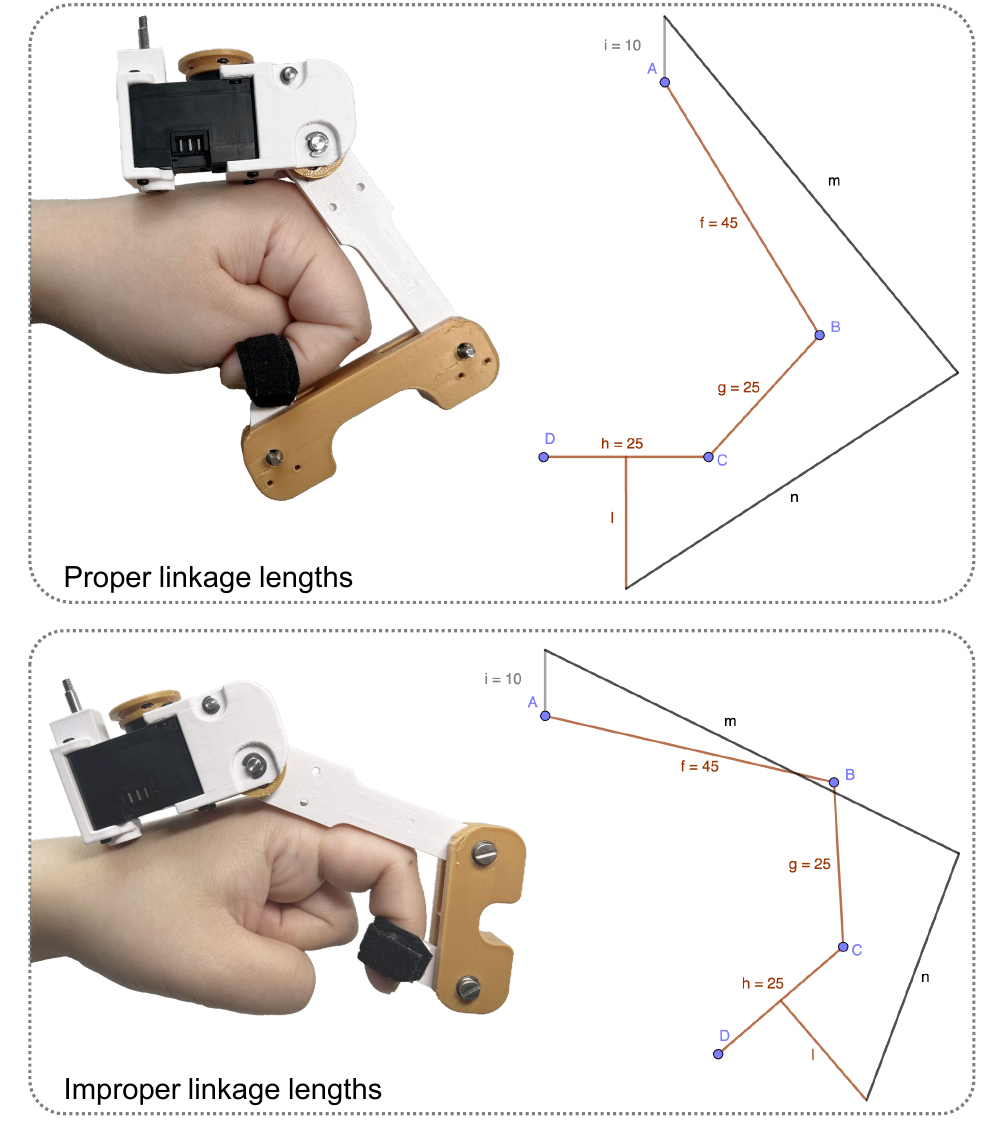}
  \caption{
  \textbf{Improper linkage lengths can cause collisions} between the human finger (\textbf{link $\boldsymbol{f, g}$}) and the glove (\textbf{link $\boldsymbol{m}$}), restricting finger movements and leading to discomfort and poor MoCap performance.
  }
  \label{fig:improper_linkage}
  \vspace{-3mm}
\end{figure}

%% file: captions/fg6-joint_explosion.tex
\begin{figure*}[t]
  \centering
  \includegraphics[width=0.95\textwidth]{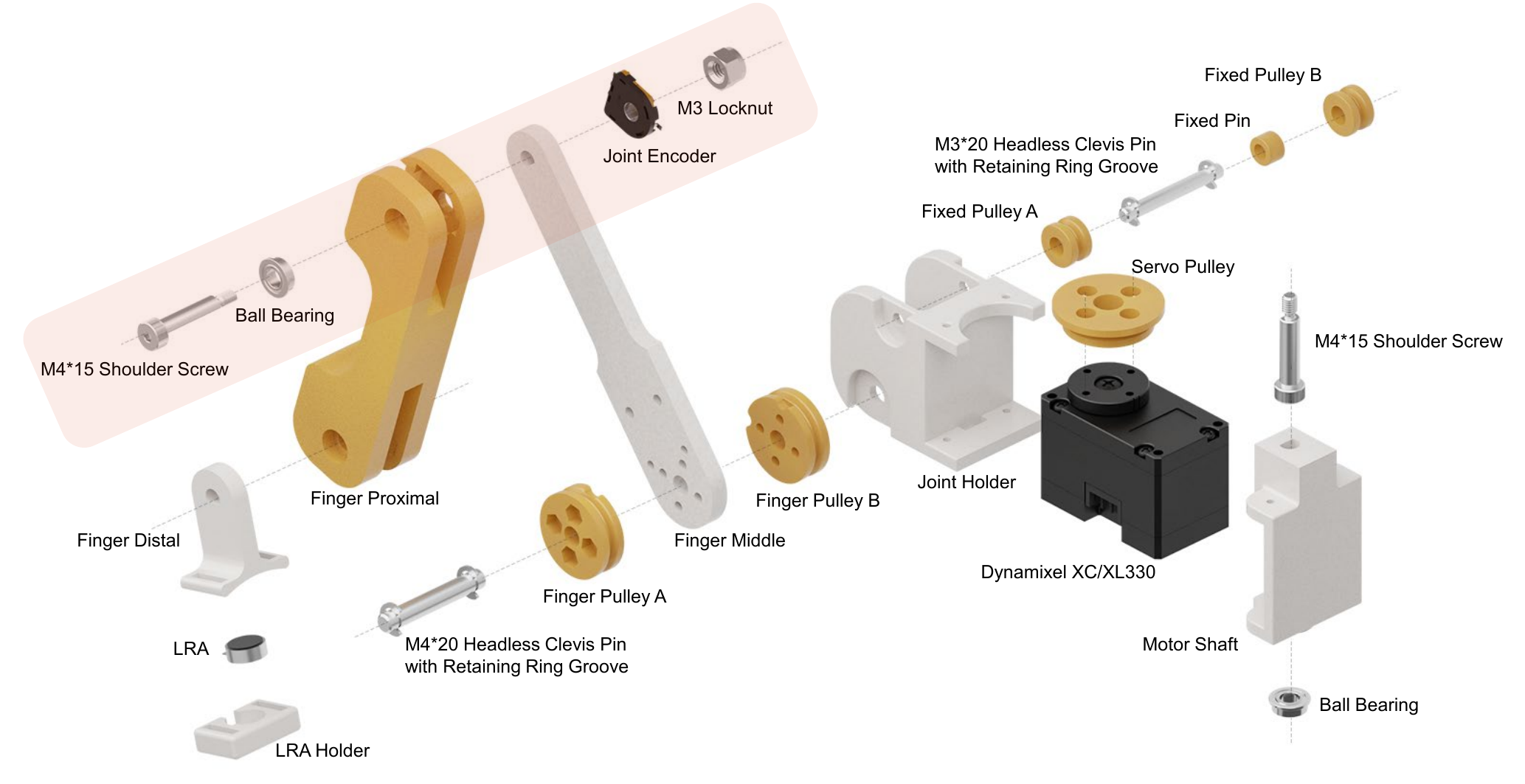}
  \caption{
  \textbf{Exploded view of the finger assembly}, with the highlighted area indicating the basic components of a rotary joint.
  }
  \label{fig:joint_explosion}
  \vspace{-3mm}
\end{figure*}

%% file: captions/fg7-pulley-system.tex
\begin{figure}[ht]
  \centering
  \includegraphics[width=0.85\linewidth]{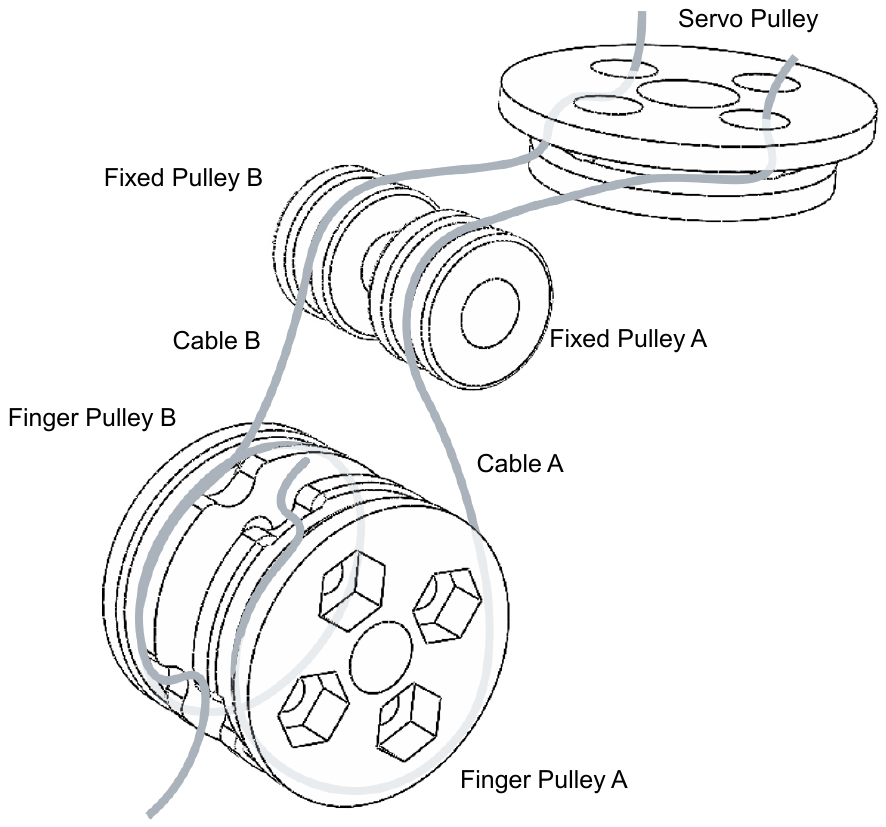}
  \caption{
  \textbf{Pulley system} of the cable-driven mechanism.
  }
  \label{fig:pulley_system}
  \vspace{-3mm}
\end{figure}

%% file: contents/5-retargeting.tex
\section{Retargeting}

\input{captions/fg8-retargeting}

\subsection{Action Retargeting}

To map human hand gestures to a robotic hand, it is essential to perform action retargeting, which converts motion data from the glove into robotic hand movements. This process addresses both the embodiment gap and motion discrepancies. 
Previous studies~\cite{wang2024dexcap, fan2023arctic, taheri2020grab} highlight the significance of fingertips, as they are the primary contact area during object interactions. Building on this insight, we apply the 5-DoF haptic force feedback to the human operators' fingertips and adopt a retargeting method focused on fingertip positions.

Our approach combines \texttt{Forward Kinematics~(FK)} to compute human fingertip positions and \texttt{Inverse Kinematics~(IK)} to calculate the corresponding robotic hand positions. 
When wearing \oursystem, the human operator secures their fingertips inside the finger caps. Since the glove acts as a rigid body, the relative positions of the fingertips with respect to the glove’s origin can be accurately calculated. With \oursystem's anthropomorphic kinematic design and precise MoCap capabilities, fingertip positions are effortlessly determined using the glove’s built-in FK.
To map these positions to a robotic hand, we utilize Mink~\cite{Zakka_Mink_Python_inverse_2024}, a differential inverse kinematics library, to generate smooth and feasible motions for the robotic hand. 

A size discrepancy often exists between the human hand and the target robotic hand. To address this, we introduce a scaling factor when calculating IK, allowing adaptation to different robotic hand sizes. This ensures an intuitive teleoperation experience where the robotic hand naturally mirrors the human hand's gestures. For instance, when the human operator opens their hand, the robotic hand open proportionally. Similarly, when the human operator brings their thumb and index finger together, the robotic hand's thumb and index finger also touch. This ability for precise fingertip alignment is critical for tasks like grasping small objects.

In our experiments, we deploy the system on the LEAP hand~\cite{shaw2023leaphand} in real-world scenarios and test it with various robotic hands in the MuJoCo simulator. Figure~\ref{fig:retargeting} presents the retargeting results of \oursystem in both simulation and real-world environments.
 
\subsection{Haptic Force Retargeting}
\label{sec::haptic_force_retargeting}
To provide the haptic force feedback, it is first necessary to sense tactile or force information at the robotic hand's fingertips. This can be achieved using a simple tactile sensor, such as the force sensing resistor~(FSR) sensor~\cite{ding2024bunny, kappassov2015tactile}, or for better performance, by utilizing an F/T sensor~\cite{kim2020six, chen2025dexforce} or a vision-based tactile sensor~\cite{yuan2017gelsight, lin20239dtact}.

In our experimental setup, we install a \texttt{1-D force sensor} on each fingertip of the LEAP Hand, with a measurement range of 3~kg and a precision of 1~g. During our quantitative experiments (Section~\ref{sec:experiments}), we identify a combination strategy for integrating haptic and force feedback that optimizes performance. This strategy along with the corresponding thresholds and feedback patterns is summarized in Table~\ref{tab:retargeting_strategy}.

\input{captions/tb1-retargeting-strategy}

When the robotic hand touches an object, the force sensor readings increase. To effectively distinguish these signal from noise, we set the first threshold at 10~g to initiate haptic feedback. During a user study without visual feedback (Section~\ref{sec::user_study}), we observe that human operators are highly sensitive to force feedback. To create a more realistic experience, force feedback is applied only after the force sensor readings exceed 50~g, which serves as the second threshold. Furthermore, observations from the bottle-slipping experiment (Section~\ref{sec::bottle-slipping}) reveal that continuous haptic feedback during teleoperation can create a misleading sensation, interfering with the operator's ability to perceive the subtle properties of the object's surface. To address this, a third threshold is set at 100~g, where haptic feedback stops, leaving only force feedback active.

For force feedback, the Dynamixel servos operate in \texttt{current-based position control mode}. The force readings from the LEAP Hand fingertips are clamped to the range [0g, 3000g], and mapped linearly to the $K_P$ gain of the Dynamixel servos. For haptic feedback, we use \texttt{waveform ID~56} from the haptic engine library, corresponding to \texttt{Pulsing Sharp 1-100\%}.

This combination strategy for haptic force retargeting enables human operators to distinguish object shape, size and softness without visual feedback. It also improves performance in complex, contact-rich manipulation tasks. Further details are provided in Section~\ref{sec:experiments}.

%% file: captions/fg8-retargeting.tex
\begin{figure*}[t]
  \centering
  \includegraphics[width=\textwidth]{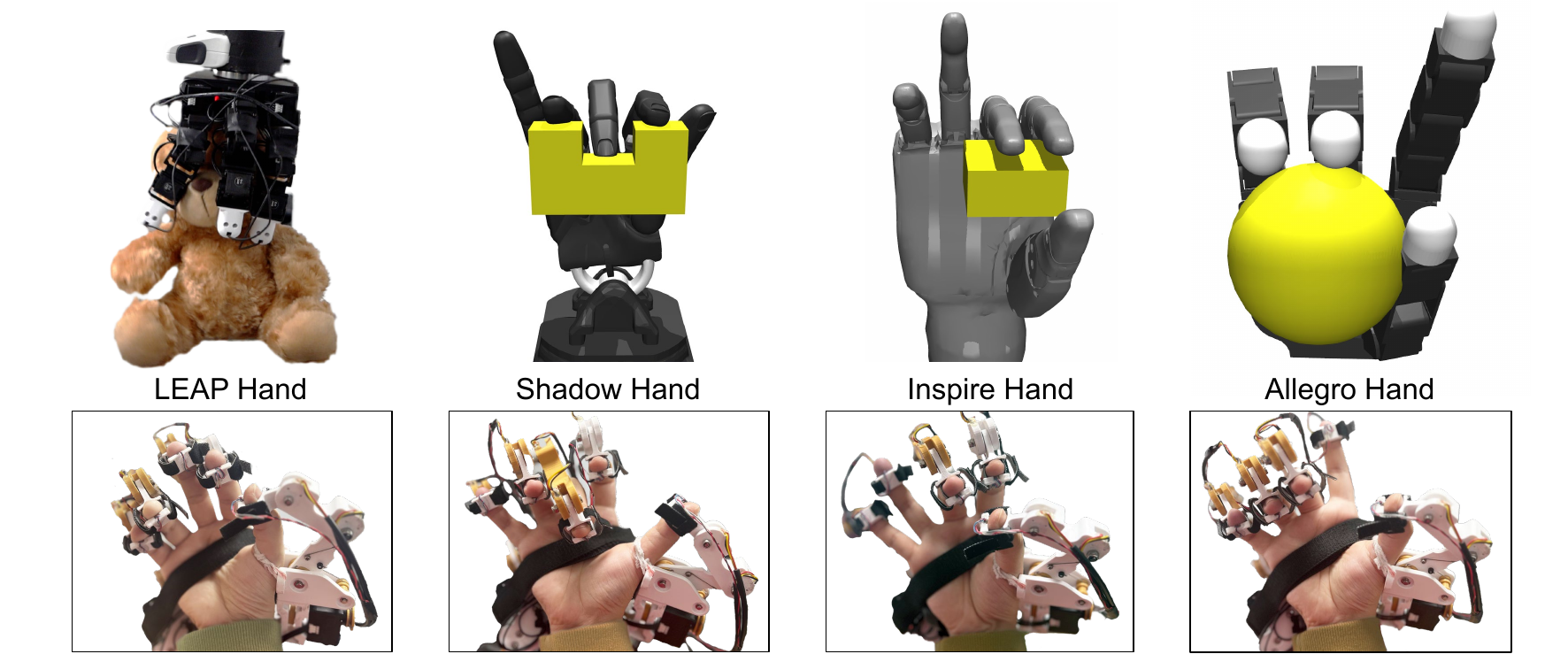}
  \caption{
  \textbf{Action retargeting results}: Teleoperating the LEAP Hand to grasp a toy in the real world and teleoperating the Shadow Hand, Inspire Hand, and Allegro Hand in simulation.
  }
  \label{fig:retargeting}
  \vspace{-5mm}
\end{figure*}

%% file: captions/tb1-retargeting-strategy.tex
\begin{table}[h]
\centering
\resizebox{0.95\linewidth}{!}{
\begin{tabular}{lcc}
\toprule
\textbf{Force Sensor Readings (g)} & \textbf{Haptic Feedback} & \textbf{Force Feedback} \\ \midrule
\textless{}10                      & \cross                  & \cross                  \\
10--50                             & \tick                   & \cross                  \\
50--100                            & \tick                   & \tick                   \\
\textgreater{}100                  & \cross                  & \tick                   \\
\bottomrule
\end{tabular}
}
\caption{\textbf{The combination strategy} for haptic force feedback in \oursystem.}
\vspace{-3mm}
\label{tab:retargeting_strategy}
\end{table}

%% file: contents/6-experiments.tex
\section{Experiments}
\label{sec:experiments}

In this section, we use \oursystem to teleoperate the \texttt{LEAP Hand}~\cite{shaw2023leaphand} mounted on the \texttt{Franka Robot Arm} to evaluate its effectiveness through a series of challenging tasks across three key aspects:

\vspace{1mm}

\hangindent=1em
\noindent\textbullet\hspace{0.5em}\textbf{Haptic Force Perception:} Without visual feedback, how effectively can \oursystem assist human operators in perceiving object properties through haptic force feedback?

\hangindent=1em
\noindent\textbullet\hspace{0.5em}\textbf{Teleoperation Efficiency:} Does integrating haptic force feedback improve vision-based teleoperation success rates and reduce task completion time? Can \oursystem enable human operators to perform challenging, contact-rich manipulation tasks?

\hangindent=1em
\noindent\textbullet\hspace{0.5em}\textbf{IL Compatibility:} Can the data collected via \oursystem be leveraged to train IL policies for dexterous manipulation?

\vspace{1mm}

\noindent \textbf{Evaluation Setup:} To evaluate the effectiveness of haptic force perception, we conduct a user study (Section~\ref{sec::user_study}) and a quantitative experiment (Section~\ref{sec::bottle-slipping}). Teleoperation efficiency is assessed in Experiment~\ref{sec::milk_box}, while IL compatibility is evaluated in Experiment~\ref{sec::IL}.

\vspace{1mm}

\noindent \textbf{Comparisons:} All experiments share the following comparison conditions, although a subset of these may be selected depending on the specific task setup:

\hangindent=1em
\noindent\textbullet\hspace{0.5em}\scaletexttt{Only Force:} Force feedback is enabled only when the force sensor readings exceed 10~g.

\hangindent=1em
\noindent\textbullet\hspace{0.5em}\scaletexttt{Only Haptic:} Haptic feedback is enabled only when the force sensor readings exceed 10~g.

\hangindent=1em
\noindent\textbullet\hspace{0.5em}\scaletexttt{Haptic+Force:} A combined feedback strategy is applied, as detailed in Section~\ref{sec::haptic_force_retargeting}.

\hangindent=1em
\noindent\textbullet\hspace{0.5em}\scaletexttt{No Haptic/Force:} \oursystem is used solely for MoCap, with no feedback provided.

\hangindent=1em
\noindent\textbullet\hspace{0.5em}\scaletexttt{Baseline:} AnyTeleop~\cite{qin2023anyteleop}, a widely recognized vision-based hand retargeting method, is used as the MoCap baseline.

\subsection{User Study: Object Perception \textbf{w/o Visual Feedback}}
\label{sec::user_study}

\input{captions/fg9-user-study}

\noindent \textbf{Task:} 
Five untrained human operators participate in this user study. During the experiment, they are required to distinguish between five pairs of objects solely through feedback from \oursystem, without any visual or auditory input (achieved by wearing an eyemask and headphones). In each trial, a pair of objects is randomly selected, and users provide their answers immediately after experiencing feedback from \oursystem for both objects.
Figure~\ref{fig:user_study} illustrates the experiment setup and the five object pairs, selected based on factors such as shape, size, and softness.

\noindent \textbf{Metrics:} 
Users' ability to distinguish object pairs is evaluated based on their success rate.

\noindent \textbf{Challenges:} 
The five object pairs are intentionally chosen based on the following considerations:

\hangindent=1em  
\noindent\textbullet\hspace{0.5em}{Pair 1:} \ul{Basic Pair}, different shape. The ball and the box have distinctly different shapes~(Fig~\ref{fig:user_study}b).

\hangindent=1em  
\noindent\textbullet\hspace{0.5em}{Pair 2:} \ul{Basic Pair}, similar shape, different size. The peanut bottle and the coffee paper cup share a similar cylindrical shape, but their diameters differ slightly~(Fig~\ref{fig:user_study}c).

\hangindent=1em  
\noindent\textbullet\hspace{0.5em}{Pair 3:} \ul{Basic Pair}, similar softness, different size. The two toys have similar softness and shapes but vary in size~(Fig~\ref{fig:user_study}d).

\hangindent=1em  
\noindent\textbullet\hspace{0.5em}{Pair 4:} \ul{Challenging Pair}, similar size and shape, different softness. Two identical bottles are used, one filled with pure water (soft) and the other filled with carbonated cola, shaken to increase its hardness~(Fig~\ref{fig:user_study}e).

\hangindent=1em  
\noindent\textbullet\hspace{0.5em}{Pair 5:} \ul{Challenging Pair}, similar shape, different size and softness. A toy cabbage (softer, larger) and a real cabbage~(Fig~\ref{fig:user_study}f).

\input{captions/tb2-user-study}

\noindent \textbf{Performance:}
As shown in Table~\ref{tab:user_study}, even without visual and auditory feedback, all participants effortlessly distinguish basic pairs 1-3.
For challenging pair 4, most participants can perceive softness using only force feedback. Some also discern softness using only haptic feedback by evaluating the duration of contact during deformation.

For challenge pair 5, when the robotic hand grasps the softer toy cabbage, it deforms to resemble the size of the real cabbage. This deformation increases its perceived softness, making it difficult for participants to distinguish using force feedback alone.

For both challenge pairs, combining haptic and force feedback slightly reduces user sensitivity, leading to a marginally lower accuracy.

\subsection{Bottle-Slipping}
\label{sec::bottle-slipping}

\input{captions/fg10-bottle}

\noindent\textit{1) Teleoperation w/o Visual Feedback}

\noindent \textbf{Task:} 
In this experiment, the human operator must perform a bottle-slipping action relying solely on feedback from \oursystem. A 15-second countdown timer is set for each trial. If the bottle successfully slips without falling within the 15 seconds, the trial is denoted as successful.

\noindent \textbf{Metrics:} 
The success rate.

\noindent \textbf{Challenges:} 
Without any visual or auditory input (achieved by wearing an eyemask and headphones), the operator must determine if the bottle is slipping at the right speed or too quickly, risking a fall.

\noindent \textbf{Performance:}
As shown in Fig~\ref{fig:bottle}a, force feedback significantly improves the success rate of this task. Additionally, incorporating haptic feedback further enhances overall performance. However, since the fingers of the LEAP Hand maintain continuous contact with the bottle during the task, haptic feedback does not provide additional information beyond using the glove solely as a MoCap device, resulting in the same success rate for both conditions.

Due to differences in retargeting strategies, even a slight change in human finger position can lead to a significant deviation in the LEAP Hand’s movements. As a result, AnyTeleop~\cite{qin2023anyteleop} struggles to perform the slipping task effectively.

\vspace{1.5mm}

\noindent\textit{2) Teleoperation with Visual Feedback}

\noindent \textbf{Task:} 
Unlike the previous blindfolded experiment, this experiment allows operators to have visual feedback. To further evaluate the operator's control ability, they are required to slip the bottle to a specified distance~(9~cm). A trial is denoted as successful if the bottle slips without falling. Additionally, We measure the deviation between the actual slipping distance and the target distance~(9~cm).

\noindent \textbf{Metrics:} 
Performance is evaluated using two metrics:

\hangindent=1em
\noindent\textbullet\hspace{0.5em}\ul{Success Rate:} A trial is denoted as successful if the bottle slips without falling.

\hangindent=1em
\noindent\textbullet\hspace{0.5em}\ul{Slipping Deviation:} This measures the difference between the target sliding distance (9~cm) and the actual slipping distance, with a smaller deviation indicating greater operational accuracy.

\noindent \textbf{Challenges:} 
Operators must precisely control the bottle to achieve the desired distance. While a greedy approach often causes the bottle to fall and results in failure, a conservative approach leads to an unsatisfactory distance deviation.

\noindent \textbf{Performance:}
This task evaluates not only success rate but also teleoperation precision. To minimize slipping deviation, operators are instructed to control the LEAP Hand carefully and optimally. As shown in Fig~\ref{fig:bottle}a, similar to previous results, haptic feedback does not provide additional information and may even interfere with task precision. However, force feedback enables operators to minimize slipping deviation more effectively. While using \oursystem solely as a MoCap device achieves the same success rate as with haptic force feedback, it results in a larger average slipping deviation.

\subsection{Rotating and Placing the Carton}
\label{sec::milk_box}

\input{captions/tb4-milk-box}
\input{captions/fg11-il}

\noindent \textbf{Task:} 
This is a long-horizon contact-rich task. As shown in Fig~\ref{fig:bottle}b, the operator must first pick up the carton horizontally, then perform an in-hand rotation, orienting the carton vertically before placing it into a small bucket.

\noindent \textbf{Metrics:} 
Performance is evaluated using two metrics:

\hangindent=1em
\noindent\textbullet\hspace{0.5em}\ul{Success Rate:} A trial is denoted as successful if the carton rotates more than 45 degrees and is successfully placed into the bucket.

\hangindent=1em
\noindent\textbullet\hspace{0.5em}\ul{Completion Time:} The total time taken to complete the entire process.

\noindent \textbf{Challenges:} 

\hangindent=1em
\noindent\textbullet\hspace{0.5em}\ul{Precise Manipulation:} The operator must accurately teleoperate to rotate the carton while preventing it from falling.

\hangindent=1em
\noindent\textbullet\hspace{0.5em}\ul{Visual Obstacle:} Grasping the carton is hindered by visual obstacles, as the operator cannot see the contact points between the robotic hand’s fingers and the carton.

\noindent \textbf{Performance:}
Table~\ref{tab:milk_box} shows that both haptic and force feedback significantly improve the teleoperation success rate and reduce completion time. While force feedback alone results in a comparable average completion time, haptic force feedback achieves a higher success rate. The vision-based MoCap method AnyTeleop~\cite{qin2023anyteleop} struggles with in-hand rotation in this task.

\subsection{Imitation Learning}
\label{sec::IL}

We show \oursystem is capable of collecting high-quality demonstrations. \textbf{3D Diffusion Policy~(DP3)}~\cite{Ze2024DP3} is selected as our imitation learning algorithm, and we use \texttt{Realsense L515} to acquire the point cloud inputs, which are then downsampled to 1024 points using farthest point sampling~\cite{qi2017pointnet}. The data collected by \oursystem is used to train policies for various downstream tasks.
We evaluate imitation learning performance on 2 basic contact-rich tasks and 1 long-horizon task: 

\noindent \textbf{Press and Move Box:} As shown in Fig~\ref{fig:il}a, the robot must continuously press down on a box and move it to a specified target location. During data collection, the box is randomly placed within a 30$\times$20 cm area, and \oursystem collects 40 demonstrations to train the policy. In evaluation, the box is also randomly placed in the same area. Across 20 trials, the success rate is 85\%~(17/20). 

\noindent \textbf{Pick and Place Teddy Bear:} As shown in Fig~\ref{fig:il}b, the robot must grasp a teddy bear and place it into a designated box. During data collection, the teddy bear’s initial position is randomized within a 30$\times$20 cm area, and \oursystem collects 40 demonstrations to train the policy. In evaluation, the bear is again randomly placed in the same area. Across 20 trials, the success rate is 70\%~(14/20), with failures primarily due to the teddy bear slipping out of the robotic hand when not grasped firmly.

\noindent \textbf{Rotating and Placing the Carton}. This task follows the same setup as Section~\ref{sec::milk_box}. For this contact-rich task, we use 3 human-collected demonstrations to train the policy. Across 10 trials, the success rate is 90\%~(9/10).

%% file: captions/fg9-user-study.tex
\begin{figure}[ht]
  \vspace{-3mm}
  \centering
  \includegraphics[width=\linewidth]{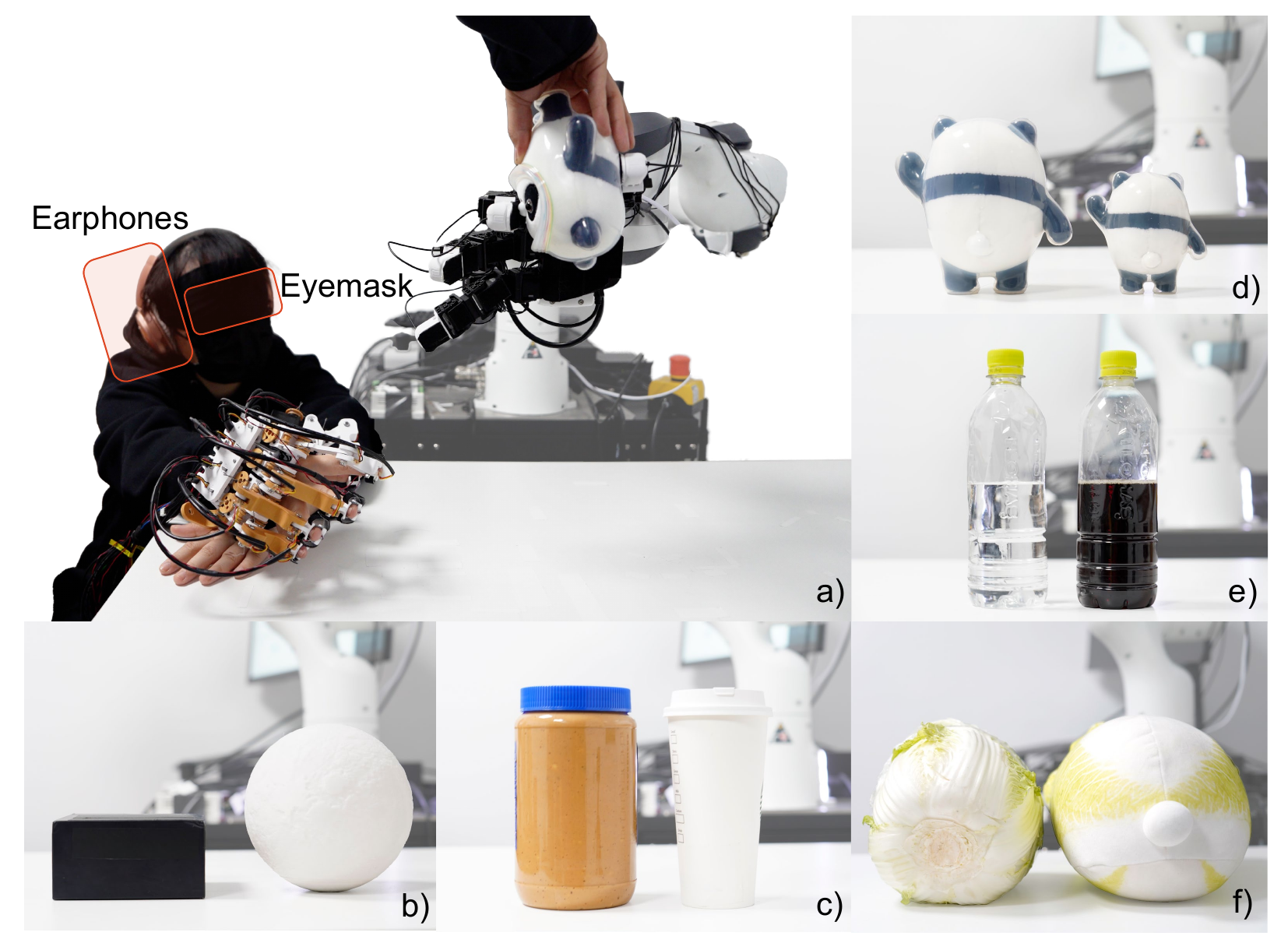}
  \caption{
  \textbf{User Study.} (a) Experiment setup: Users wear an eyemask and headphones to eliminate visual and auditory feedback. (b)–(f) Object pairs tested in the study.
  }
  \label{fig:user_study}
  \vspace{-3mm}
\end{figure}

%% file: captions/tb2-user-study.tex
\begin{table}[h]
\centering
\resizebox{0.95\linewidth}{!}{
\begin{tabular}{l|ccccc}
    \toprule
    & \textbf{Pair 1} & \textbf{Pair 2} & \textbf{Pair 3} & \textbf{Pair 4} & \textbf{Pair 5} \\ 
    \midrule
    \textbf{Only Force}   & 5/5 & 5/5 & 5/5 & \textbf{4/5} & 0/5 \\ 
    \textbf{Only Haptic}  & 5/5 & 5/5 & 5/5 & 3/5 & \textbf{3/5} \\ 
    \textbf{Haptic+Force} & 5/5 & 5/5 & 5/5 & 3/5 & 2/5 \\ 
    \bottomrule
\end{tabular}
}
\caption{\textbf{Success rates in the user study.} All feedback modes perform well for the basic pairs. For the challenging pairs, force feedback is more sensitive to softness, while haptic feedback is more sensitive to shape.}
\vspace{-3mm}
\label{tab:user_study}
\end{table}

%% file: captions/fg10-bottle.tex
\begin{figure*}[t]
  \vspace{-3mm}
  \centering
  \includegraphics[width=\textwidth]{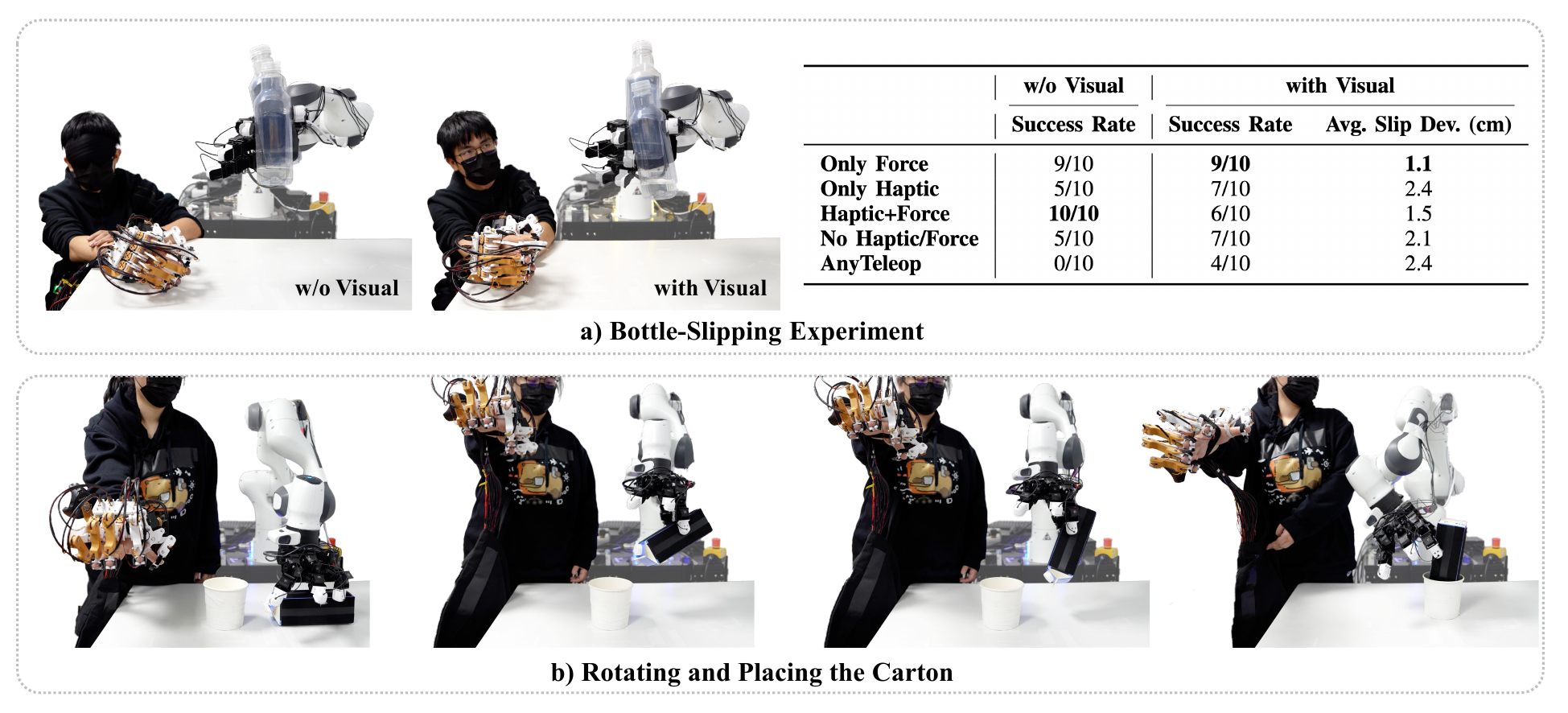}
  \caption{
  \textbf{Teleoperation experiments and quantitative results.}
  a) Without visual feedback, force feedback significantly improves the task success rate. With visual feedback, it enhances precise control. 
  b) In in-hand rotation, the challenge is to slightly release the fingers, allowing the carton to rotate without slipping out (as shown in the middle two images).
  }
  \label{fig:bottle}
  \vspace{-5mm}
\end{figure*}

%% file: captions/tb4-milk-box.tex
\begin{table}[h]
    \vspace{-1mm}
    \centering
    \resizebox{0.95\linewidth}{!}{
       \begin{tabular}{lcc}
            \toprule
            & \textbf{Success Rate} & \textbf{Average Completion Time (s)} \\
            \midrule
            \textbf{Only Force} & 9/10 & \textbf{18.92} \\
            \textbf{Only Haptic} & 9/10 & 21.16 \\
            \textbf{Haptic+Force} & \textbf{10/10} & \textbf{19.89} \\
            \textbf{No Haptic/Force} & 4/10 & 24.76 \\
            \textbf{AnyTeleop} & 1/10 & 54.85 \\
            \bottomrule
        \end{tabular}
    }
    \caption{\textbf{Quantitative experiment results.} Haptic force feedback enables operators to achieve a higher success rate and a faster average completion time, as haptic feedback provides contact information, while force feedback indicates the proper timing for in-hand rotation.}
    \vspace{-3mm}
    \label{tab:milk_box}
\end{table}

%% file: captions/fg11-il.tex
\begin{figure*}[t]
  \vspace{-3mm}
  \centering
  \includegraphics[width=\textwidth]{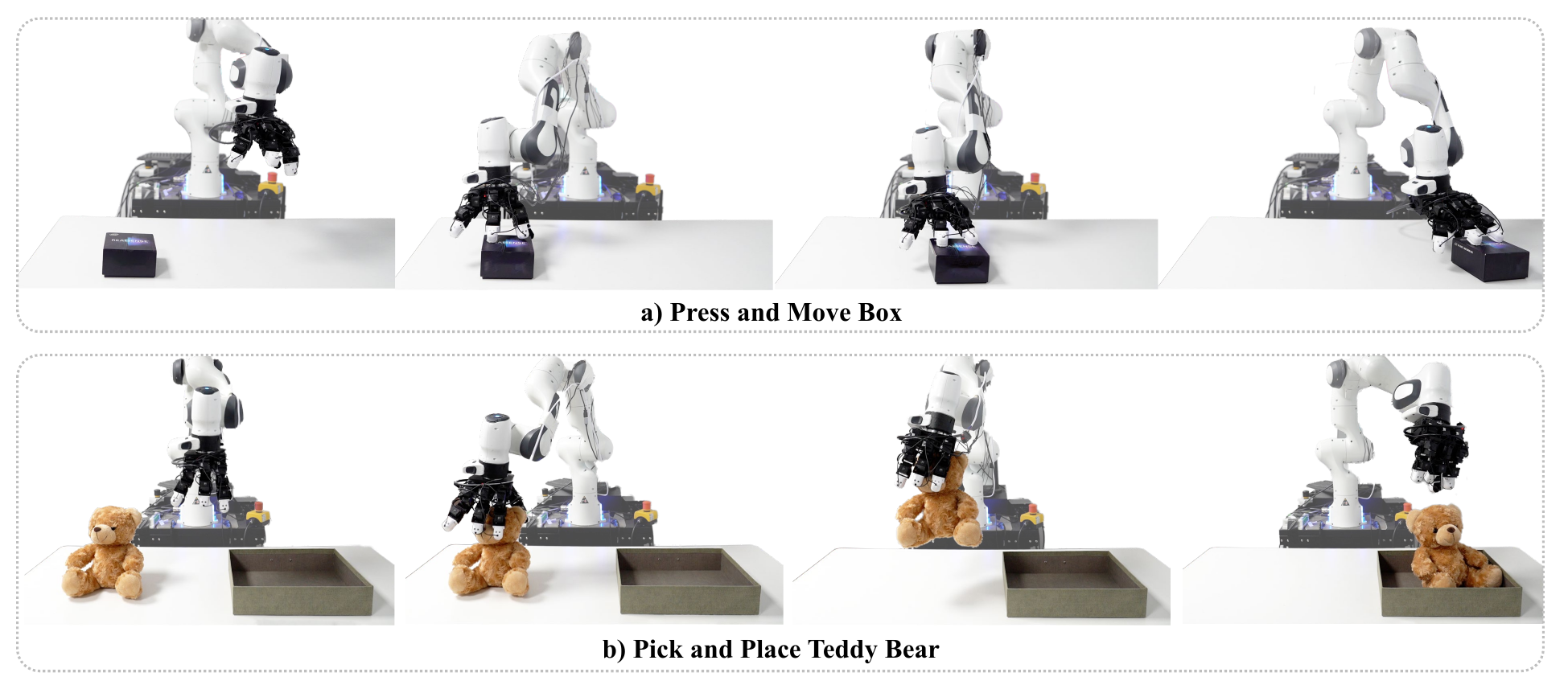}
  \caption{
  \textbf{The imitation learning experiment.}
  (a) The robot must first locate the correct position of the box and then apply adequate force to press it. Excessive force prevents movement, while insufficient force causes the fingers to slip.
  (b) The robot must first locate the bear, then open its hand to grasp it. Due to the bear's size, precise grasping control is required. An inaccurate grasp deforms the bear and causes it to slip out of the fingertips.
  }
  \label{fig:il}
  \vspace{-5mm}
\end{figure*}

%% file: contents/7-limitations.tex
\section{Limitations and Future Work}

\oursystem is a powerful haptic force feedback glove for dexterous manipulation, but several limitations remain.
First, the weight of \oursystem is inevitably high, as it utilizes 5 commercial servos, bringing the total weight to 550g. Additionally, in agile teleoperation scenarios, performance is constrained by the servos' maximum speed and torque output.
To address these issues, we are investigating the use of lighter servos with smaller reduction ratios and designing a customized reduction mechanism to balance speed and torque more effectively.
Second, although \oursystem is designed to accommodate most hand sizes, it may be uncomfortable for some users. 
To enhance adaptability and wearability, we are developing CAD files for linkages in multiple sizes, enabling customization for various hand dimensions.

%% file: contents/8-conclusion.tex
\section{Conclusion} 
\label{sec:conclusion}
In this paper, we present \oursystem, a low-cost, open-source haptic force feedback glove designed for dexterous manipulation. \oursystem enables precise and efficient execution of long-horizon, contact-rich tasks. Experimental results show that \oursystem enhances the operator's immersive teleoperation experience while also serving as an effective tool for training imitation learning policies. Moreover, the user study demonstrates that \oursystem provides precise perception of object properties through its integrated haptic force feedback. To support further research and contribute to the community, all hardware designs and code will be open-sourced.

%% file: contents/x-acknowledgment.tex
\section*{Acknowledgment}
We would like to thank Zhengrong Xue, Gu Zhang, Changyi Lin, Mengda Xu, and Yifan Hou for their invaluable advice and fruitful discussions on hardware design and learning policies.
We also appreciate Wenhao Ding and Laixi Shi for their insightful discussions and feedback.
Additionally, we thank Yichuan Gao, Xiaoyan Yang, Xinyao Qin, and Botian Xu for their assistance with the user study.
Special thanks to Skyentific, Gennady Plyushchev, for their innovative contributions to the unconventional cable-driven joint design.
We are also grateful to Tiansheng Sun and Guanghan Pan for their open-source repository of the HTC Vive Tracker Python API.
Lastly, we extend our appreciation to Yitong Wang for her help in creating elegant graphic renderings of the hardware design.